\documentclass{article}

\usepackage[main, preprint, nonatbib]{neurips_2026}

\usepackage[utf8]{inputenc} 
\usepackage[T1]{fontenc}    
\usepackage{hyperref}       
\hypersetup{
colorlinks=true,
linkcolor=blue, 
citecolor=cyan
}
\usepackage{placeins}
\usepackage{url}            
\usepackage{booktabs}       
\usepackage{amsfonts}       
\usepackage{nicefrac}       
\usepackage{microtype}      
\usepackage{xcolor}         
\usepackage{graphicx}
\usepackage{multirow}
\usepackage{enumitem}
\usepackage[table]{xcolor}
\usepackage{xcolor}
\usepackage{wrapfig}
\usepackage{amsmath,amssymb}

\newcommand{\ie}{\emph{i.e.}}
\title{Diagnosing and Correcting Bias in MLLM \\ for Long Video Understanding}

\author{%
  \textbf{Xusheng Liang}$^{1,2}$ \quad
  \textbf{Jianqiao Sun}$^{3,4}$\thanks{Contribution partially done while Jianqiao Sun was an intern at JX.} \quad
  \textbf{Hao Zhang}$^{3}$ \quad
  \textbf{Yulei Niu} \\
  \textbf{Hengshuang Zhao}$^{5,2}$ \quad
  \textbf{Jiawei Ma}$^{1}$
  \\
  $^{1}$City University of Hong Kong \quad
  $^{2}$Shenzhen Loop Area Institute \quad
  $^{3}$Xidian University
  \\
  $^{4}$JX Intelligence Technology Co., Ltd. \quad
  $^{5}$The University of Hong Kong
  \\
  \texttt{xushliang2-c@my.cityu.edu.hk} \quad
  \texttt{jiaweima@cityu.edu.hk}
}

\begin{document}

\maketitle

\begingroup
\renewcommand{\thefootnote}{$\dagger$}
\endgroup



{

\begin{abstract}
Long-video question answering is challenging since the answer often hinges on a few decisive moments scattered throughout the video, while memory constraints force Multimodal Large Language Models (MLLMs) to sample frames at extremely low rates, resulting in sparsity that risks omitting key evidence. Though aggressively enlarging the context size is intuitive, we argue that performance remains fundamentally constrained by the bias, \emph{i.e.}, the misleading cues caused by spurious \textit{linguistic} correlations and salient yet irrelevant \textit{visual} observations. In this paper, we first introduce {S-MME}, a benchmark designed to systematically diagnose shortcut-biased behavior in MLLMs. To mitigate this failure, we further propose a training-free framework, Counterfactual Long-video Evidence-Aware Reasoning (CLEAR). By treating the full video-question pair as a complete view, CLEAR estimates bias effects via augmented views by only keeping the question or locally salient visual clips. Then, we intervene on hidden states to mitigate the discrepancy at inference such that the correct answer can be predicted confidently. With experimental studies across five benchmarks, we show that our method is generalizable and can be integrated in different MLLMs, yielding consistent gains in prediction and robustness. We hope CLEAR can offer a complementary direction in test-time scaling for long-video understanding and will publish code.
\end{abstract}

\section{Introduction}

Visual understanding over a long temporal horizon is inherently challenging because the correct answer often hinges on a few decisive frames, \ie, \emph{evidence}, sparsely dispersed across the video. Then, memory constraints force current Multimodal Large Language Models (MLLMs) to sample frames at extremely low rates as the input context, further compounding this difficulty and inducing the model to rely on spurious correlations, \ie, \emph{shortcuts}, for prediction. To mitigate predictions, the intuitive methods are to scale up the sampling rate and context window~\cite{zhang2024long}, increase visual token capacity~\cite{wang2024qwen2}, and strengthen instruction tuning~\cite{chen2024sharegpt4video}. Nevertheless, they all incur substantially higher computational cost and inevitably face an efficiency–effectiveness trade-off.

At the same time, enlarging the context alone is insufficient, and performance remains fundamentally constrained by biases even when relevant frames are completely fed into the model. 
As illustrated in Fig.~\ref{fig:intro_diagnosis}, the model is prone to follow misleading cues, including (left) spurious linguistic correlations~\cite{agrawal2018don}, where the phrasing of the question and answer options makes a distractor plausible without viewing the video, and (right) salient yet causally irrelevant observations, \ie, frames that are semantically related to the query yet do not determine the answer. Similarly, while existing methods such as frame selection~\cite{yu2024frame}, token pruning~\cite{shen2024longvu} and memory mechanisms~\cite{song2024moviechat}, aim to filter out irrelevant frames before feeding them into the model, they still assume that the model can be robust enough to use the key observation. To this end, we believe that evaluation should go beyond aggregate benchmark scores and systematically target samples that are sensitive to shortcuts.


\begin{figure}[t]
    \centering
    \includegraphics[width=\linewidth]{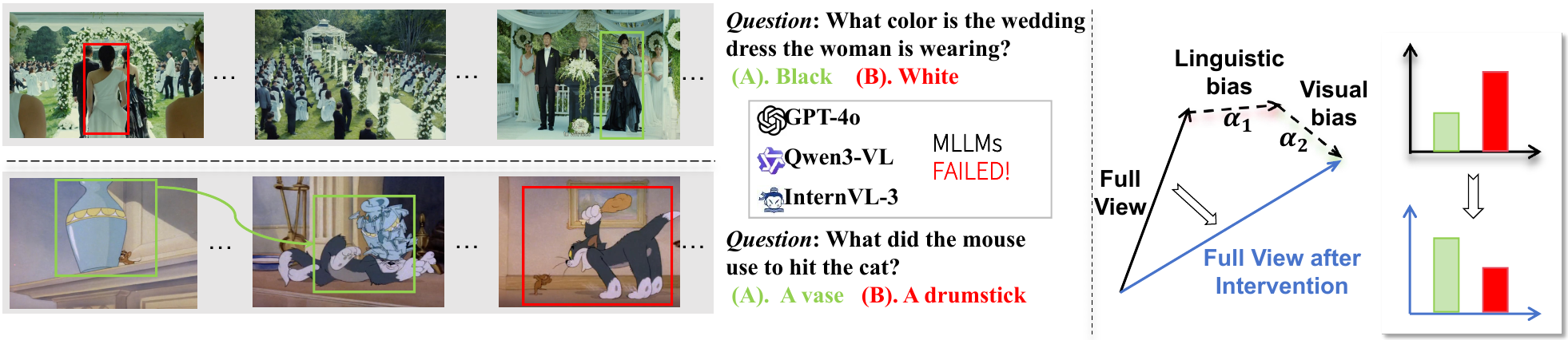}
    \caption{MLLMs often fail despite the presence of decisive evidence, either by following spurious linguistic correlations or by over-relying on salient but misleading visual observations.
CLEAR diagnoses these shortcut-biased trajectories with controlled probes and steers the full-video representation toward evidence-consistent prediction.
    }\label{fig:intro_diagnosis}
\end{figure}
In detail, we construct S-MME, standing for \textit{Shortcuts in Multi-Modal Evaluation}, a long-video benchmark composed of video-question pairs designed to expose shortcut-sensitive failures, \ie, when given a restricted observation that forces a spurious correlation, models produce an incorrect prediction, whereas access to the evidence frames can yield the correct answer. Then, following the examples in Fig.~\ref{fig:intro_diagnosis} and adopting the concept of counterfactual samples~\cite{niu2021counterfactual}, we construct the restricted view in two aspects: a question-only view where the question and options alone make distracted prediction, and a salient-yet-irrelevant view where individual clips have high semantic similarity to the question in a common representation space yet lack the decisive evidence required for the correct answer. In this way, we can directly test whether a model follows the decisive clips or instead collapses to a counterfactual alternative.

To further mitigate the bias to such spurious shortcut, we propose a training-free framework Counterfactual Long-video Evidence-Aware Reasoning (CLEAR). CLEAR runs the frozen model, treats the full video-question pair as a complete view, and constructs counterfactual views that maximally preserve the shortcut signals and weaken or even remove decisive evidence. Specifically, we use a question-only view to expose linguistic bias and a salient view formed from highly similar visual observations to expose visual bias. We posit that the prediction discrepancy between views serves as an estimate of the bias effect, and we apply lightweight residual steering in deep layers to intervene on hidden states and suppress spurious correlations, thereby minimizing the distraction and enabling the model to predict the correct answer with more confidence. Meanwhile, our method is generalizable and can be integrated into different MLLMs without additional training, and yields consistent gains through comprehensive experimental studies. Finally, our contributions are summarized as follows:

\begin{itemize}[leftmargin=*,itemsep=0pt]
    \item We identify bias from 1) spurious linguistic correlations and 2) salient but misleading visual observations as a fundamental bottleneck in long-video understanding by MLLMs, which persists beyond sparse frame sampling alone.
    \item To examine such bias of existing MLLMs, we introduce S-MME, a benchmark for systematically diagnosing this behavior. The S-MME considers spurious correlation caused by either linguistic correlation or salient-yet-misleading views.
    \item We propose CLEAR, a training-free framework that estimates bias effects from counterfactual views and mitigates them through hidden-state intervention, thereby encouraging predictions to rely on decisive evidence. The method is generalizable and can be applied to different MLLMs operating in either classification or next-token prediction modes.
    \item Through experimental study, we demonstrate consistent improvements over three public datasets, Video-MME~\cite{fu2025video}, LongVideoBench~\cite{wu2024longvideobench}, and MLVU~\cite{zhou2024mlvu}, as well as our S-MME benchmark. 
    

    

\end{itemize}

\section{Related Works}

\paragraph{Multimodal LLMs for Long-Video Understanding} Recent Video-LLMs improve long-video understanding by increasing the amount of visual evidence that can be processed within limited model context. 
Existing methods typically compress frame-level features into compact visual tokens~\cite{liu2025oryx,zohar2025apollo,ye2024mplug,wang2025videollamb,liu2025video}, reduce spatial-temporal redundancy through pooling, token aggregation, or adaptive filtering~\cite{maaz2024video,xu2024slowfast,xu2024pllava,jin2024video}, or extend multimodal context length to support more frames and longer reasoning horizons~\cite{zhang2024long,wang2024qwen2,team2023gemini,chen2025eagle, ma2023digeo}. 
While these methods improve evidence access, often with non-trivial memory or computational cost~\cite{li2024videochat,shu2025video,li2024knn, li2025vtfsl, li2026dvla}, our work studies a complementary problem: whether accessible evidence is faithfully used when shortcut cues provide plausible answers.

\paragraph{Shortcut Learning and Bias in Visual Question Answering} Shortcut learning has long been recognized as a central challenge in visual question answering. 
Early studies showed that VQA models can exploit answer priors, question templates, or dataset correlations without sufficient visual grounding~\cite{agrawal2018don,cadene2019rubi, li2026semmsa, liang2026multimodal,pu2026f}. 
Subsequent work exposed and mitigated such biases through distribution-shifted benchmarks, language-bias suppression, stronger visual grounding, and counterfactual sample construction~\cite{agrawal2018don,clark2019don,chen2020counterfactual,niu2021counterfactual}. 
While these studies provide important insights, they mainly focus on static-image VQA or conventional multimodal fusion. 
Long-video reasoning introduces an additional shortcut source: a locally salient clip may appear predictive but miss temporally distributed events, state changes, or later evidence required for the correct answer. 
Recent video QA benchmarks also suggest that final answer accuracy alone does not fully reveal whether models ground their predictions in the relevant temporal evidence~\cite{xiao2021next,wu2024longvideobench, ma2024mode}. 
Our work extends shortcut diagnosis from language-only bias to both language-prior and partial-observation shortcuts in long-video reasoning.

\paragraph{Training-Free Methods for Long-Video Understanding} Training-free long-video methods improve frozen Video-LLMs through inference-time input selection, memory construction, or agentic reasoning, without updating model parameters~\cite{liu2025towards}. 
Selection-based methods reduce redundant visual input through adaptive keyframe sampling, token merging, or query-guided retrieval~\cite{tang2025adaptive,zhang2025beyond,yu2024frame,zhang2025flexselect}. 
Memory-enhanced methods maintain long-range information using event memory, temporal caches, recursive summarization, or tree-structured video representations~\cite{song2024moviechat,cheng2025enhancing,santos2025video,wang2025videotree}. 
Agent-based methods formulate long-video QA as an active inference process, iteratively retrieving segments, revising plans, and performing multi-step reasoning with language agents~\cite{wang2024videoagent,chen2025lvagent,zhi2025videoagent2,wang2026videochat,wu2024surgbox}. 
These approaches mainly optimize which evidence is selected, retained, or queried during inference.

\section{S-MME: Shortcuts in Multi-Modal Evaluation}
\label{sec:smme}

\subsection{Making Evidence Use Diagnosable}

Long-video QA evaluation typically reports whether a model produces the correct answer. 
However, final-answer accuracy alone does not reveal whether the model has relied on the decisive evidence.
A model may fail because the decisive evidence is missing from the sampled context, but it may also fail even when such evidence is available, because the prediction is dominated by shortcut cues such as spurious linguistic correlations or salient-yet-misleading visual observations.
We refer to the latter as biased evidence use: the model observes relevant evidence but still follows a plausible shortcut. This behavior is difficult to isolate with long-video benchmarks, where a wrong answer may arise from missing frames, temporal ambiguity, or shortcut-induced bias.
To make evidence use diagnosable, we construct cases in which complete video evidence and restricted shortcut cues support competing answers.
If a model predicts the shortcut-consistent distractor under the full-video input, the failure indicates that the model does not fully use the decisive evidence. Accordingly, we introduce S-MME, a diagnostic benchmark for evaluating shortcut-robust long-video reasoning.

\subsection{Shortcut-Sensitive Cases}

The basic unit of S-MME is a shortcut-sensitive long-video QA case.
Each case consists of a long video, a question, answer candidates, a sufficient evidence span, and an annotated shortcut-consistent distractor. 
A valid conflict case must satisfy three conditions. 
First, the full video contains sufficient evidence to determine a unique correct answer. 
Second, at least one restricted observation, such as the question alone or a short salient clip, is insufficient for determining the correct answer. 
Third, this restricted view makes a specific incorrect answer plausible, yielding a shortcut-consistent distractor.

We consider two dominant shortcut sources. The first comes from spurious linguistic correlations, where the question or answer options alone make an incorrect answer plausible without visual grounding.
The second comes from salient-yet-misleading visual observations, where a visually salient segment is semantically related to the question but does not determine the answer without broader temporal context. These two sources reflect common shortcut pathways in long-video QA and provide counterfactual views for diagnosis. Beyond assigning each case to one shortcut source, S-MME annotates the sufficient evidence span and the shortcut-consistent distractor for every sample.


\begin{figure}[t]
\centering
\includegraphics[width=\linewidth]{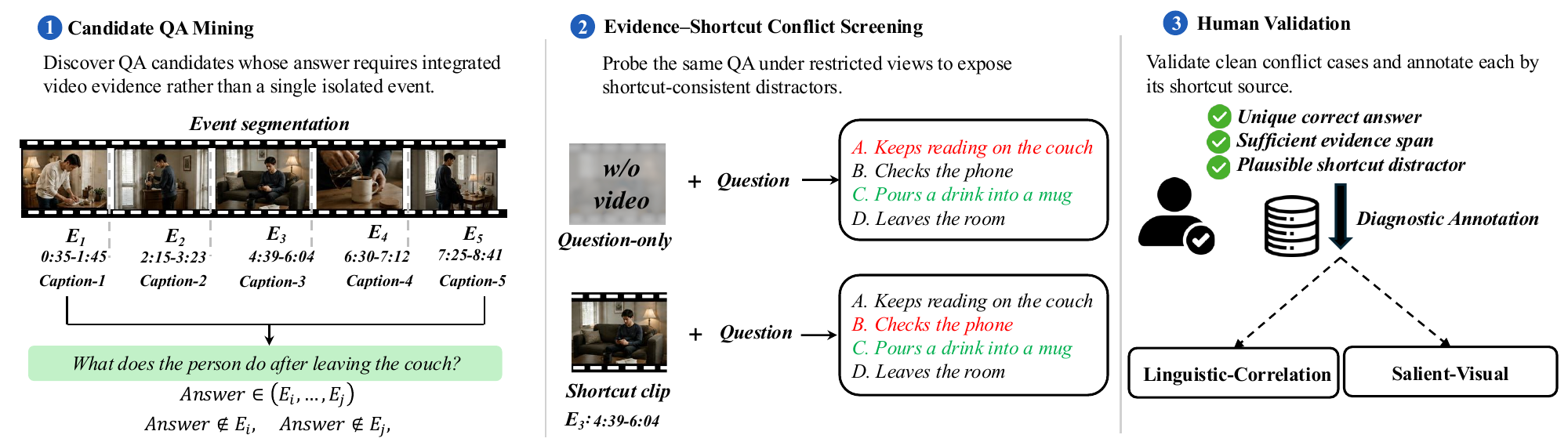}
\caption{
\textbf{Construction pipeline of S-MME.}
We first mine candidate questions with explicit long-video evidence requirements, then use restricted views to identify shortcut-consistent distractors, followed by human validation of answer uniqueness, evidence sufficiency, and shortcut plausibility.
}
\label{fig:benchmark_pipeline}
\end{figure}



\begin{table}[t]
\centering
\caption{Statistics of the S-MME diagnostic benchmark.}
\label{tab:s-mme-stats}
\small 
\begin{tabular}{lcccc}
\toprule
\textbf{Category} & \textbf{Count}  & \textbf{Avg. Dur. (min)} & \textbf{Min Dur. (min)} & \textbf{Max Dur. (min)} \\ \midrule
Linguistic-Correlation Conflict & 148  & 11.9 & 5.6 & 74.7 \\
Salient-Visual Conflict & 172 & 17.6 & 12.8 & 106.6 \\ \midrule
\textbf{Total} & \textbf{320}  & \textbf{15.0} & \textbf{5.6} & \textbf{106.6} \\ \bottomrule
\end{tabular}
\end{table}

\subsection{Benchmark Construction}

We construct S-MME through a three-stage pipeline, as illustrated in Figure~\ref{fig:benchmark_pipeline}. 
The first stage mines candidate QA samples with explicit long-video evidence requirements. 
Given a long video, we segment it into coherent temporal events and summarize the timeline with concise event descriptions. 
Candidate questions are generated from these event summaries to require evidence beyond a single isolated observation, such as cross-event comparison, temporal state change, delayed outcome, or local observation contradicted by later evidence. 
For each candidate, annotators mark the sufficient evidence span and verify that the full video uniquely determines the correct answer.

The second stage uses restricted views to identify shortcut-consistent distractors.
For the question-only view, annotators check whether the question and answer options alone make a particular incorrect option plausible without visual grounding.
For the salient-view, annotators identify a short segment that is semantically related to the question and supports a plausible distractor, but does not contain sufficient evidence to determine the correct answer without broader video context.
These restricted views are used only for candidate discovery and do not determine the ground truth of a sample.

The third stage applies human validation and quality control.
Annotators retain a candidate only if it satisfies three criteria:
(i) the full video uniquely supports the correct answer,
(ii) the restricted view is insufficient for determining the correct answer, and
(iii) the annotated shortcut-consistent distractor is plausible under the restricted view.
Ambiguous questions, multiple-answer cases, noisy videos, and samples without a clearly identifiable evidence span are removed.
This meticulous validation pipeline ensures that S-MME evaluates whether models can leverage available evidence to override plausible shortcuts, effectively decoupling reasoning robustness from annotation noise or missing information.

\section{CLEAR: Counterfactual Long-video Evidence-Aware Reasoning}
\label{sec:method}

\begin{figure}[t]
    \centering
    \includegraphics[width=\linewidth]{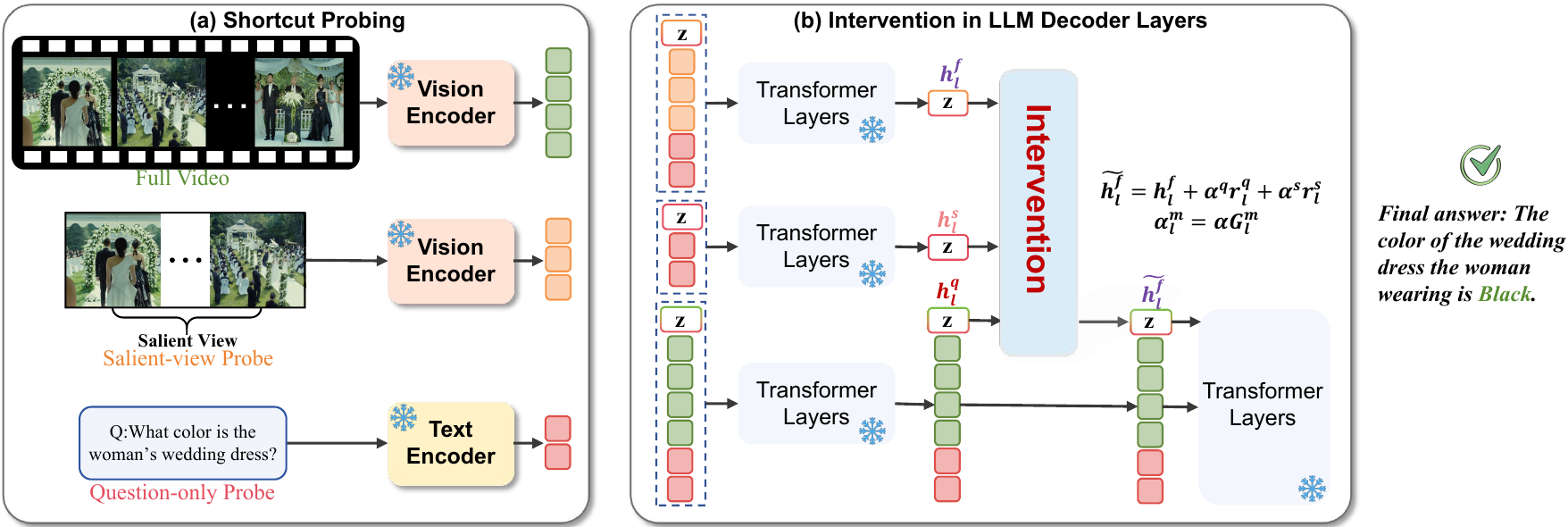}
\caption{
\textbf{Overview of CLEAR.}
(a) The full-video path produces the final answer, while question-only and salient-view probes reveal spurious linguistic correlations and salient-yet-misleading visual observations. The salient view is formed by selecting top-$K$ locally relevant clips according to query-conditioned visual attention.
(b) CLEAR extracts the hidden state of the shared decision token \(z\) at selected late decoder layers, and constructs residual vectors \(r_l^q = h_l^f - h_l^q\) and \(r_l^s = h_l^f - h_l^s\) from the restricted-view probes to the full-video state.
These residual vectors are injected into the full-video path to steer its hidden state away from shortcut anchors before final decoding.
}
    \label{fig:method_overview}
\end{figure}

\subsection{Problem Setup}
\label{sec:problem_setup}

We consider a video MLLM parameterized by $\theta$, which takes a video
$v$ and a textual query $x$ as input, and generates a response
$y=(y_1,\ldots,y_T)$ in an auto-regressive manner:
\begin{equation}
y_t \sim p_\theta(y_t \mid v, x, y_{<t}),
\end{equation}
where $y_{<t}$ denotes previously generated tokens.

Long-video understanding requires the model to integrate evidence
across temporally distant clips. Errors may still occur when the
relevant evidence is present, because the model can form a reasoning trajectory
dominated by biases. CLEAR addresses this issue by exposing such shortcut trajectories
through counterfactual views and correcting the full-video trajectory through
residual steering. 

\subsection{Shortcut Probing via Counterfactual Views}
\label{sec:shortcut_views}

To make shortcut-biased reasoning observable, CLEAR evaluates the same
video-question sample under counterfactual observations. These observations are not
used to produce the final answer. Instead, they provide diagnostic trajectories
that reveal how the model behaves when complete video evidence is removed or
restricted. We focus on two common shortcut sources in long-video reasoning:
language priors from the question itself and local visual shortcuts from
salient but incomplete clips.

\vspace{0.15cm}
\noindent\textbf{Question-only probe.}
The question-only probe removes the video input and keeps only the textual
query. This view captures the model's spurious linguistic correlations, allowing the model
to predict without visual evidence. For
questions containing semantic associations or stereotypical
object-action relations, the model may assign high probability to a plausible
but visually unsupported answer. 

\vspace{0.15cm}
\noindent\textbf{Salient-view probe.}
The salient-view probe keeps only a small number of locally relevant clips.
Given a video $v$, we uniformly divide it into $N$ non-overlapping temporal
clips $\{c_i\}_{i=1}^{N}$, where each clip contains consecutive frames rather
than an isolated frame. We then compute a query-conditioned relevance score
$a_i$ for each clip and select the top-$K$ clips:


\begin{equation}
\mathcal{I}_{\mathrm{sal}}
=
\operatorname{TopK}_{i\in\{1,\ldots,N\}}(a_i),
\qquad
v^s = \{c_i \mid i \in \mathcal{I}_{\mathrm{sal}}\},
\label{eq:salient_view}
\end{equation}
where $\mathcal{I}_{\mathrm{sal}}$ denotes the indices of the selected clips and $v^s$ denotes the resulting salient-view input.

In our implementation, the clip-level score is computed from the model's
attention response from the query tokens to the visual tokens of each clip:
\begin{equation}
    a_i
    =
    \frac{1}{|\mathcal{Q}|\,|\mathcal{T}_i|}
    \sum_{q \in \mathcal{Q}}
    \sum_{t \in \mathcal{T}_i}
    A_{q \rightarrow t},
    \label{eq:clip_attention_score}
\end{equation}
where $Q$ denotes the set of query tokens, $T_i$ denotes the visual tokens belonging to clip $c_i$, and $A_{q\to t}$ denotes the attention response from query token $q$ to visual token $t$. We use this score to select candidate clips for the salient-view probe.

The salient view is not intended to retrieve the complete evidence. Instead,
it approximates the model's most immediately attended local context. Such local
clips may contain objects, actions, or scenes that are highly related to the
question, but they may still miss the temporal context required for the correct
answer. Therefore, this probe exposes visually prominent but incomplete evidence.

To compare the full-video path with the restricted-view probes, we extract
hidden states at the same textual decision position. For each view, we run a
forward pass with the same textual query $x$ and extract the hidden state of
the shared decision token $z$ before answer generation:
\begin{equation}
    \mathbf{h}_l^m
    =
    H_l(v_m,x)[z],
    \qquad
    m\in\{f,q,s\},
    \label{eq:hidden_state_definition}
\end{equation}
where $H_l(\cdot)$ denotes the hidden states at the $l$-th LLM decoder layer, 
and $z$ is the last textual token before answer generation. 
We use the same prompt template across all views and extract $\mathbf{h}_l^f$, 
$\mathbf{h}_l^q$, and $\mathbf{h}_l^s$ at this fixed decision position, ensuring 
that only the visual observations differ.

\subsection{Residual Contrastive Steering}
\label{sec:residual_steering}

The counterfactual probes above expose alternative trajectories formed under
incomplete observations. CLEAR uses these trajectories as contrastive anchors
to correct the full-video reasoning process. The key intuition is that the
difference between the full-video trajectory and a shortcut-probe trajectory
defines a direction from restricted-view reasoning toward complete-observation
reasoning.

\vspace{0.15cm}
\noindent\textbf{Shortcut residuals.}
Given the full-video hidden state $\mathbf{h}_l^f$ and a shortcut-probe hidden
state $\mathbf{h}_l^m$ at layer $l$, where $m\in\{q,s\}$, we define:
\begin{equation}
\mathbf{r}_l^m = \mathbf{h}_l^f - \mathbf{h}_l^m .
\end{equation}
This residual is not assumed to be a disentangled shortcut component. Instead,
it is a directional contrast between incomplete-observation reasoning and
full-video reasoning. In other words, it reflects the representational contrast
between the restricted-view trajectory and the full-video trajectory.

For the two counterfactual probes,
$\mathbf{r}_l^q$ provides a correction direction against  spurious linguistic correlations, while $\mathbf{r}_l^s$ provides a correction direction against
salient-yet-misleading visual observations.



\noindent\textbf{Shortcut-repulsive update.}
CLEAR updates the full-video hidden state by injecting residual corrections
from the activated shortcut probes:
\begin{equation}
\tilde{\mathbf{h}}_l^f
=
\mathbf{h}_l^f
+
\alpha_l^q \mathbf{r}_l^q
+
\alpha_l^s \mathbf{r}_l^s ,
\label{eq:shortcut_update}
\end{equation}
where the effective intervention strength for each probe is defined as
\begin{equation}
\alpha_l^m = \alpha G_l^m,\qquad
G_l^m = \mathbb{I}(d_l^m > \tau),\qquad
m\in\{q,s\}.
\label{eq:effective_alpha}
\end{equation}
Here, $\alpha$ is a global steering coefficient, $\tau$ is a fixed discrepancy threshold, and $G_l^m$ is a binary gate indicating whether the shortcut-probe trajectory sufficiently deviates from the full-video trajectory at layer $l$. Thus, a residual is injected only when the corresponding shortcut probe is activated.

Since each residual is defined as
$\mathbf{r}_l^m=\mathbf{h}_l^f-\mathbf{h}_l^m$, it points from the
restricted-view shortcut anchor to the full-video state. Adding activated
residuals therefore moves the full-video state away from the shortcut anchors.
The remaining decoder layers then continue computation from the corrected state.

\begin{wrapfigure}[13]{r}{0.45\linewidth}
    \centering
    \vspace{0pt}
    \includegraphics[width=0.98\linewidth]{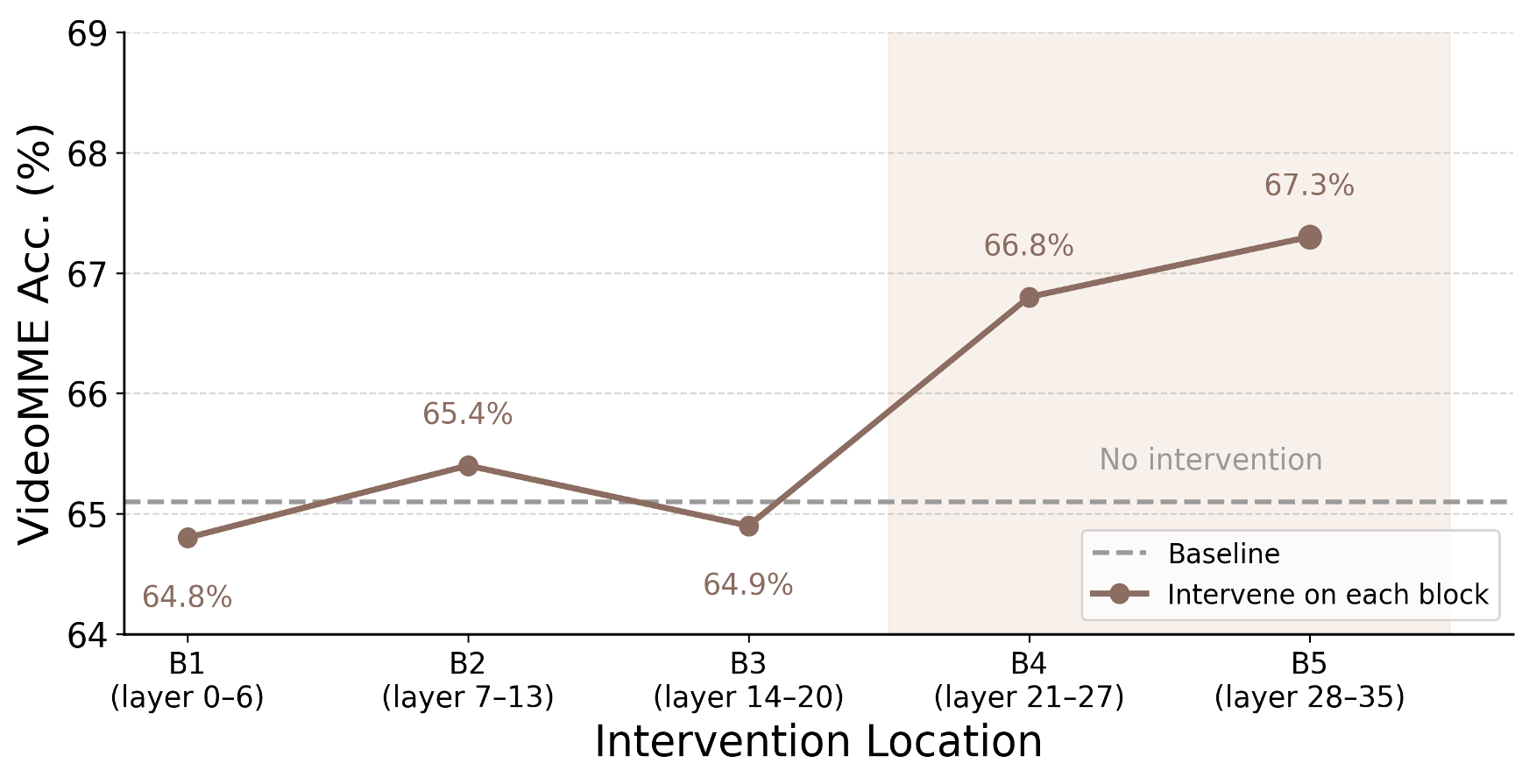}
    \caption{
        \textbf{Effect of intervention location.}
        Late-layer intervention performs best,
        with the last block achieving the highest accuracy.
    }
    \label{fig:block_analysis}
    \vspace{0pt}
\end{wrapfigure}

\paragraph{Where and when to intervene.}
Although residual directions can be constructed at every layer, not all layers are equally suitable for correction.

We empirically compare intervention at different decoder blocks and find that late-layer intervention is the most effective, with the last block, layers 28--35, giving the best performance, as shown in Figure~\ref{fig:block_analysis}.
This is consistent with the intuition that answer-oriented representations are formed near the decoding stage.

Within the selected layers, we further determine when to apply intervention based on the discrepancy between the full-video trajectory and the shortcut-probe trajectory:
\begin{equation}
d_l^m = 1 - \cos(\mathbf{h}_l^f, \mathbf{h}_l^m).
\end{equation}
A larger $d_l^m$ indicates that the shortcut probe induces a more distinct reasoning trajectory.
We therefore activate intervention only when $d_l^m > \tau$, using a fixed threshold $\tau=0.15$ in all experiments.

\section{Experiments}
\label{sec:experiments}

\definecolor{lightgray}{RGB}{235,235,235}
\definecolor{goodgreen}{RGB}{46,125,50}
\newcommand{\cleargain}[1]{\textcolor{goodgreen}{\scriptsize$\uparrow$#1}}
\newcommand{\clearreduce}[1]{\textcolor{goodgreen}{\scriptsize$\downarrow$#1}}

\subsection{Experimental Setup}
\label{sec:exp_setup}

\begin{table*}[t]
\centering
\scriptsize
\setlength{\tabcolsep}{3.0pt}
\renewcommand{\arraystretch}{0.90}
\caption{
\textbf{General comparison on long-video understanding benchmarks.}
CLEAR is evaluated on three open-source MLLMs, with closed-source and public models included for reference.
}

\label{tab:main_general}
\resizebox{\textwidth}{!}{
\begin{tabular}{l c c c c c c c}
\toprule
Model
& Size
& Frames
& MLVU
& \multicolumn{2}{c}{Video-MME}
& LongVB
& LVBench \\
\cmidrule(lr){5-6}
& & & 
& Overall & Long 
& & \\
\midrule
\multicolumn{8}{l}{\textit{Closed-source MLLMs}} \\
GPT-4o~\cite{gpt4o}
& --
& --
& 54.9
& 65.3
& 59.0
& \textbf{66.7}
& 48.9 \\
Gemini-1.5-Pro~\cite{team2023gemini}
& --
& --
& 53.8
& \textbf{67.4}
& \textbf{61.8}
& 64.0
& 33.1 \\

\midrule
\multicolumn{8}{l}{\textit{Open-source video MLLMs}} \\
LongVU~\cite{shen2024longvu}
& 7B
& 1 fps$^\dagger$
& 63.7
& 60.9
& 54.3
& 59.5
& -- \\

NVILA~\cite{liu2025nvila}
& 8B
& 8
& 57.7
& 64.2
& 54.8
& 57.7
& -- \\

VideoLLaMA3~\cite{zhang2025videollama}
& 7B
& 1 fps$^\dagger$
& 63.0
& 66.2
& --
& 59.8
& 45.3 \\

mPLUG-Owl3~\cite{ye2024mplug}
& 7B
& 256
& 62.1
& 64.3
& 53.0
& 58.7
& 32.6 \\

ViLAMP~\cite{cheng2025scaling}
& 7B
& 1 fps$^\dagger$
& 62.3
& 67.3
& 59.3
& 62.0
& 30.8 \\

VideoChatFlash~\cite{li2024videochat}
& 7B
& 16
& 64.7
& 65.3
& 55.4
& 64.7
& 48.2 \\

\midrule
LLaVA-Video~\cite{zhang2024llava}
& 7B
& 64
& 52.1
& 59.0
& 47.2
& 58.6
& 43.1 \\
\quad + CLEAR
& 7B
& 64
& 55.3 \cleargain{3.2}
& 61.5 \cleargain{2.5}
& 49.8 \cleargain{2.6}
& 61.0 \cleargain{2.4}
& 49.0 \cleargain{5.9} \\

\midrule
InternVL3~\cite{zhu2025internvl3}
& 8B
& 64
& 57.8
& 63.1
& 52.7
& 59.4
& 43.4 \\
\quad + CLEAR
& 8B
& 64
& 62.4 \cleargain{4.6}
& 66.7 \cleargain{3.6}
& 56.9 \cleargain{4.2}
& 62.3 \cleargain{2.9}
& 49.7 \cleargain{6.3} \\

\midrule
Qwen3-VL~\cite{bai2025qwen3}
& 8B
& 64
& 59.0
& 65.1
& 55.5
& 64.3
& 46.0 \\
\quad + CLEAR
& 8B
& 64
& \textbf{66.8} \cleargain{7.8}
& \textbf{70.9} \cleargain{5.8}
& \textbf{62.7} \cleargain{7.2}
& \textbf{68.1} \cleargain{3.8}
& \textbf{52.4} \cleargain{6.4} \\
\bottomrule
\end{tabular}
}
\par\noindent
\makebox[\textwidth][l]{\normalfont\scriptsize $^\dagger$~1 fps: sampling one frame per second over the full video.}

\end{table*}

\paragraph{Models and Benchmarks.}
We evaluate CLEAR on representative open-source long-video MLLMs, including LLaVA-Video-7B, InternVL3-8B, and Qwen3-VL-8B. Closed-source models, including GPT-4o and Gemini-1.5-Pro, are reported only as reference systems since their hidden states are inaccessible. Experiments are conducted on four public long-video benchmarks, Video-MME, LongVideoBench, LVBench, and MLVU, covering general video QA, long-context temporal reasoning, and complex event-level understanding. We further evaluate on S-MME, our diagnostic benchmark for measuring whether models collapse to shortcut-consistent alternatives.

\subsection{Main Results on Public Long-video Benchmarks}
\label{sec:public_results}

\paragraph{Compared to open-source baselines.}
As shown in Table~\ref{tab:main_general}, CLEAR consistently improves
open-source long-video MLLMs under the same final-frame budget and decoding
setting. The gains are most pronounced on reasoning-intensive benchmarks:
CLEAR improves Qwen3-VL-8B by +7.8 on MLVU and +7.2 on Video-MME-Long. For InternVL3-8B, CLEAR improves Video-MME Overall by +3.6, Video-MME Long by +4.2, and LVBench by +6.3. On LLaVA-Video-7B, CLEAR also
achieves a clear +5.9 improvement on LVBench. 

These results show that CLEAR improves general long-video understanding without
increasing the final-frame budget or changing the backbone. The stronger gains
on MLVU and Video-MME-Long suggest that standard long-video benchmarks also
contain evidence-use failures, where relevant evidence is present but models
still rely on shortcut cues. Closed-source models are included as references,
while CLEAR is applied to open-source models with accessible hidden states.

\begin{table*}[t]
\centering
\scriptsize
\setlength{\tabcolsep}{2.0pt}
\renewcommand{\arraystretch}{0.88}
\caption{
\textbf{Diagnostic comparison on S-MME.} ShortErr denotes the percentage of predictions that incorrectly choose the annotated shortcut-consistent distractor.
}
\label{tab:smme_diagnostic}
\resizebox{\textwidth}{!}{
\begin{tabular}{l c c cc cc cc}
\toprule
Model
& Size
& Frames
& \multicolumn{2}{c}{Lang.-Prior}
& \multicolumn{2}{c}{Part.-Evid.}
& \multicolumn{2}{c}{Overall} \\
\cmidrule(lr){4-5}\cmidrule(lr){6-7}\cmidrule(lr){8-9}
& & 
& Acc.$\uparrow$ & ShortErr.$\downarrow$
& Acc.$\uparrow$ & ShortErr.$\downarrow$
& Acc.$\uparrow$ & ShortErr.$\downarrow$ \\
\midrule
\multicolumn{9}{l}{\textit{Human Reference}} \\
Human
& --
& --
& 92.4 & 2.6
& 90.1 & 3.8
& 91.2 & 3.2 \\

\midrule
\multicolumn{9}{l}{\textit{Closed-source MLLMs}} \\
GPT-4o
& --
& --
& 47.2 & 18.6
& 43.8 & 22.4
& 45.5 & 20.5 \\

Gemini-1.5-Pro
& --
& --
& 54.1 & 14.2
& 50.3 & 17.9
& 52.2 & 16.1 \\

\midrule
\multicolumn{9}{l}{\textit{Open-source video MLLMs }} \\
LongVU
& 7B
& 1 fps$^\dagger$
& 15.6 & 51.8
& 12.4 & 56.7
& 14.0 & 54.3 \\

NVILA
& 8B
& 8
& 17.9 & 48.6
& 14.8 & 53.1
& 16.4 & 50.9 \\

VideoLLaMA3
& 7B
& 1 fps$^\dagger$
& 18.6 & 47.5
& 15.3 & 52.4
& 17.0 & 49.9 \\

mPLUG-Owl3
& 7B
& 256
& 19.1 & 46.8
& 16.0 & 51.6
& 17.6 & 49.2 \\

ViLAMP
& 7B
& 1 fps$^\dagger$
& 18.3 & 47.9
& 15.7 & 52.0
& 17.0 & 50.0 \\

VideoChatFlash
& 7B
& 16
& 16.8 & 50.2
& 13.9 & 54.6
& 15.4 & 52.4 \\
\midrule
LLaVA-Video
& 7B
& 64
& 11.1 & 56.8
& 7.9 & 61.4
& 9.5 & 59.1 \\
\quad + CLEAR
& 7B
& 64
& 14.8 \cleargain{3.7}
& 51.7 \clearreduce{5.1}
& 10.6 \cleargain{2.7}
& 56.2 \clearreduce{5.2}
& 12.7 \cleargain{3.2}
& 53.9 \clearreduce{5.2} \\

\midrule
InternVL3
& 8B
& 64
& 19.8 & 44.3
& 16.5 & 48.0
& 18.2 & 46.2 \\
\quad + CLEAR
& 8B
& 64
& 24.6 \cleargain{4.8}
& 38.5 \clearreduce{5.8}
& 18.9 \cleargain{2.4}
& 42.4 \clearreduce{5.6}
& 21.7 \cleargain{3.5}
& 40.5 \clearreduce{5.7} \\

\midrule
Qwen3-VL
& 8B
& 64
& 22.2 & 41.7
& 18.8 & 46.2
& 20.5 & 44.0 \\
\quad + CLEAR
& 8B
& 64
& \textbf{28.4} \cleargain{6.2}
& \textbf{34.1} \clearreduce{7.6}
& \textbf{24.5} \cleargain{5.7}
& \textbf{39.6} \clearreduce{6.6}
& \textbf{26.5} \cleargain{6.0}
& \textbf{36.9} \clearreduce{7.1} \\
\bottomrule
\end{tabular}
}
\par\noindent
\makebox[\textwidth][l]{\normalfont\scriptsize $^\dagger$~1 fps: sampling one frame per second over the full video.}

\end{table*}

\subsection{Diagnostic Evaluation on S-MME}
\label{sec:smme_results}

Table~\ref{tab:smme_diagnostic} shows that S-MME exposes substantial
evidence-use failures in current long-video MLLMs. Human annotators reach
91.2\% accuracy with only 3.2\% ShortErr, while the strongest closed-source
model achieves 52.2\% accuracy and still produces 16.1\% shortcut-consistent
errors. Open-source models exhibit more severe shortcut collapse, with the best
baseline Qwen3-VL reaching only 20.5\% accuracy and 44.0\% ShortErr. CLEAR improves both accuracy and shortcut robustness. On Qwen3-VL, it raises
overall accuracy from 20.5\% to 26.5\% and reduces ShortErr from 44.0\% to
36.9\%. The reduction appears in both conflict types, suggesting that the two counterfactual probes address complementary
shortcut sources.

\begin{table}[t]
\centering
\small
\setlength{\tabcolsep}{3.5pt}
\renewcommand{\arraystretch}{0.90}
\caption{
\textbf{Effect of shortcut probes on S-MME.}
Q-only and S-only apply residual correction from the language-prior and salient-view probes, respectively. 
CLEAR combines both probes for complementary correction.
}
\label{tab:probe_ablation}
\begin{tabular}{l c c l cc cc c}
\toprule
Variant & Q & S & Branch
& \multicolumn{2}{c}{Linguistic Bias}
& \multicolumn{2}{c}{Visual Bias}
& Overall \\
\cmidrule(lr){5-6}\cmidrule(lr){7-8}
& & &
& Acc.$\uparrow$ & ShortErr.$\downarrow$
& Acc.$\uparrow$ & ShortErr.$\downarrow$
& Acc.$\uparrow$ \\
\midrule
Baseline & -- & -- & --
& 22.2 & 41.7
& 18.8 & 46.2
& 20.5 \\
Q-only   & \checkmark & -- & Single
& 26.1 & 36.8
& 20.4 & 43.5
& 23.3 \\
S-only   & -- & \checkmark & Single
& 24.5 & 38.7
& 22.9 & 41.0
& 23.7 \\
\midrule
\rowcolor{gray!8}
CLEAR & \checkmark & \checkmark & Dual
& \textbf{28.4} & \textbf{34.1}
& \textbf{24.5} & \textbf{39.6}
& \textbf{26.5} \\
\bottomrule
\end{tabular}
\end{table}

\begin{table*}[t]
\centering
\scriptsize
\renewcommand{\arraystretch}{0.90}
\setlength{\tabcolsep}{3.5pt}

\caption{
\textbf{Efficiency and intervention analysis.}
(a) On Video-MME, CLEAR outperforms simple frame scaling under comparable inference budgets.
(b) On S-MME, moderate residual strength yields the best accuracy--shortcut suppression trade-off.
}
\label{tab:efficiency_intervention}

\begin{minipage}[t]{0.56\textwidth}
\centering
\parbox[t][2.2em][c]{\linewidth}{\centering \textbf{(a) Efficiency on Video-MME}}

\begin{tabular*}{\linewidth}{@{\extracolsep{\fill}} l c c c c}
\toprule
Method & Final F. & Aux. Probe & Time & Acc.$\uparrow$ \\
\midrule
Qwen3-VL & 32  & --  & 0.62$\times$ & 62.8 \\
Qwen3-VL & 64  & --  & 1.00$\times$ & 65.1 \\
Qwen3-VL & 128 & --  & 1.86$\times$ & 67.0 \\
\midrule
Qwen3-VL + CLEAR & 32  & Q+S & 1.34$\times$ & 68.4 \cleargain{5.6} \\
Qwen3-VL + CLEAR & 64  & Q+S & 1.78$\times$ & \textbf{70.9} \cleargain{5.8} \\
Qwen3-VL + CLEAR & 128 & Q+S & 2.72$\times$ & 71.5 \cleargain{4.5} \\
\bottomrule
\end{tabular*}
\end{minipage}
\hfill
\begin{minipage}[t]{0.40\textwidth}
\centering
\parbox[t][2.2em][c]{\linewidth}{\centering \textbf{(b) Intervention strength on S-MME}}

\begin{tabular*}{\linewidth}{@{\extracolsep{\fill}} c c c c}
\toprule
Method & $\alpha$ & Acc.$\uparrow$ & ShortErr.$\downarrow$ \\
\midrule
Qwen3-VL & 0.00 & 20.5 & 44.0 \\
\midrule
\multicolumn{1}{c}{\multirow{5}{*}{\begin{tabular}[c]{@{}c@{}}Qwen3-VL\\ + CLEAR\end{tabular}}}
& 0.05 & 23.4 \cleargain{2.9} & 40.6 \clearreduce{3.4} \\
& 0.10 & 25.1 \cleargain{4.6} & 38.5 \clearreduce{5.5} \\
& \cellcolor{lightgray}0.15 
& \cellcolor{lightgray}\textbf{26.5} \cleargain{6.0} 
& \cellcolor{lightgray}36.9 \clearreduce{7.1} \\
& 0.25 & 26.0 \cleargain{5.5} & 36.5 \clearreduce{7.5} \\
& 0.35 & 24.8 \cleargain{4.3} & \textbf{36.1} \clearreduce{7.9} \\
\bottomrule
\end{tabular*}
\end{minipage}
\end{table*}

\subsection{Is More Context Enough?}
\label{sec:context_scaling}
A natural way to improve long-video understanding is to sample more frames.
We compare the baseline and CLEAR under different frame budgets to test whether context scaling alone resolves shortcut-biased errors.
On S-MME, we report both accuracy and ShortErr; on Video-MME, we report accuracy as a representative public benchmark. Figure~\ref{fig:context_scaling} shows that more frames improve baseline accuracy, confirming the importance of evidence access.
However, ShortErr decreases more slowly, indicating that context scaling alone does not fully resolve biased evidence use.
Across all frame budgets, CLEAR achieves higher accuracy and lower ShortErr, and its gains on Video-MME suggest that shortcut correction generalizes beyond S-MME.

\begin{figure}[t]
    \centering
    \includegraphics[width=0.99\linewidth]{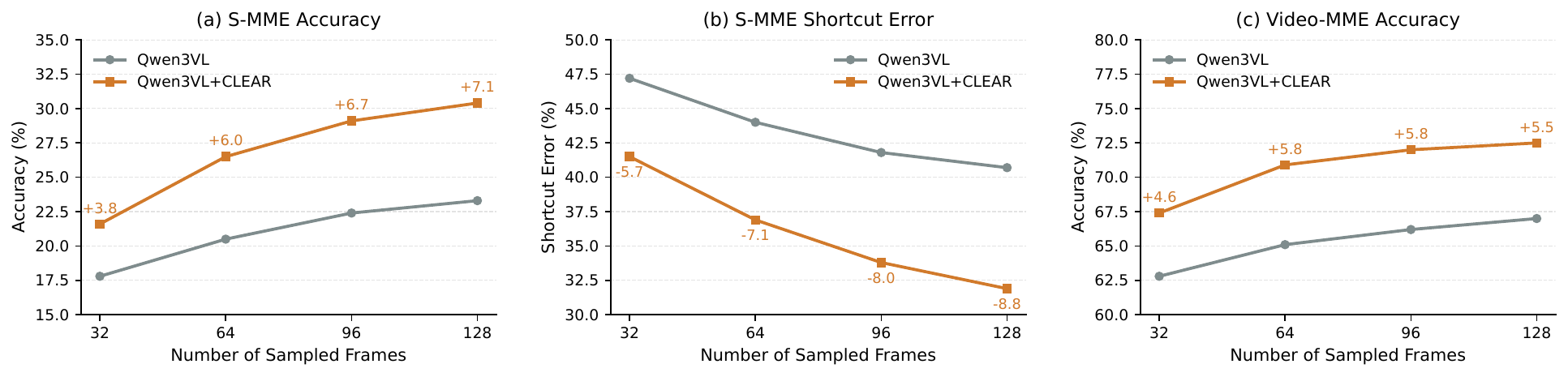}
    \caption{
    \textbf{Context scaling analysis.}
CLEAR consistently improves accuracy and reduces ShortErr under the same frame budget.
    }
    \label{fig:context_scaling}
\end{figure}

\subsection{Do Shortcut Probes Align with Conflict Types?}
\label{sec:probe_ablation}

CLEAR uses two restricted views to expose different bias sources: the question-only probe targets spurious linguistic correlations, while the salient-view probe targets salient-yet-misleading visual observations.
We compare single-branch and dual-branch variants on S-MME to examine whether these probes play source-specific roles.
Table~\ref{tab:probe_ablation} shows that Q-only mainly improves cases dominated by spurious linguistic correlations, indicating its ability to capture question-option priors.
S-only is more effective on cases involving salient-yet-misleading visual observations, suggesting that locally plausible visual cues cannot be diagnosed from language alone.
Dual-branch variants further improve overall robustness, showing that the two restricted views provide complementary correction signals.

\subsection{Efficiency and Intervention Analysis}
\label{sec:efficiency_intervention}

Although CLEAR introduces auxiliary shortcut probes, its goal is not to simply increase inference compute.
We compare it with a direct compute-scaling baseline that uses more full-video frames.
Table~\ref{tab:efficiency_intervention}(a) shows that increasing the full-video context brings limited gains with noticeably higher cost, while CLEAR achieves larger accuracy improvements under comparable inference budgets.
This suggests that the benefit mainly comes from shortcut diagnosis and residual corrections.

We further study the strength of residual correction.
Table~\ref{tab:efficiency_intervention}(b) shows that a moderate residual scale gives the best accuracy--robustness trade-off.
When $\alpha$ becomes too large, ShortErr continues to decrease slightly, but accuracy starts to drop, indicating over-correction.
Overall, CLEAR remains training-free: the backbone is frozen, and the probes only provide inference-time correction signals.

\section{Conclusion}

We studied shortcut-driven failures in long-video MLLMs, where models may rely
on spurious linguistic correlations or salient yet misleading visual observations
despite the presence of decisive evidence. To diagnose this behavior, we
introduced S-MME, a benchmark that tests whether models follow complete video
evidence or collapse to shortcut-consistent alternatives. We further proposed
CLEAR, a training-free inference-time framework that uses counterfactual views to
expose shortcut-induced trajectories and applies late-layer residual steering
to the full-video path. Experiments on public long-video benchmarks and S-MME
show that CLEAR consistently improves accuracy and reduces shortcut-consistent
errors across different MLLMs.


\section*{Acknowledgments and Disclosures}

Jiawei Ma acknowledges the support of the National Natural Science Foundation of China (NSFC) under Grant No.~62502405 and the Hong Kong Research Grants Council (RGC) under Grant No.~TRS T43-101/26-N. This work was also supported by the National Natural Science Foundation of China (NSFC) under Grant Nos.~62422606 and 62441615, the Hong Kong Research Grants Council (RGC) General Research Fund (GRF) under Grant No.~17213925, and the Shenzhen Loop Area Institute Fund under Grant No.~FPF10120250006. Hao Zhang acknowledges the support of the National Natural Science Foundation of China (NSFC) under Grant No.~62301384 and the Excellent Young Scientists Fund (Overseas).

\bibliographystyle{plain}
\bibliography{references}

@inproceedings{liu2025oryx,
  title={{Oryx MLLM}: On-Demand Spatial-Temporal Understanding at Arbitrary Resolution},
  author={Liu, Zuyan and Dong, Yuhao and Liu, Ziwei and Hu, Winston and Lu, Jiwen and Rao, Yongming},
  booktitle={ICLR},
  year={2025}
}

@inproceedings{zohar2025apollo,
  title={Apollo: An exploration of video understanding in large multimodal models},
  author={Zohar, Orr and Wang, Xiaohan and Dubois, Yann and Mehta, Nikhil and Xiao, Tong and Hansen-Estruch, Philippe and Yu, Licheng and Wang, Xiaofang and Juefei-Xu, Felix and Zhang, Ning and others},
  booktitle={Proceedings of the Computer Vision and Pattern Recognition Conference},
  pages={18891--18901},
  year={2025}
}

@inproceedings{wang2025videollamb,
  title={Videollamb: Long streaming video understanding with recurrent memory bridges},
  author={Wang, Yuxuan and Song, Yiqi and Xie, Cihang and Liu, Yang and Zheng, Zilong},
  booktitle={Proceedings of the IEEE/CVF International Conference on Computer Vision},
  pages={24170--24181},
  year={2025}
}

@inproceedings{liu2025video,
  title={Video compression commander: Plug-and-play inference acceleration for video large language models},
  author={Liu, Xuyang and Wang, Yiyu and Ma, Junpeng and Zhang, Linfeng},
  booktitle={Proceedings of the 2025 Conference on Empirical Methods in Natural Language Processing},
  pages={1910--1924},
  year={2025}
}

@inproceedings{song2024moviechat,
  title={Moviechat: From dense token to sparse memory for long video understanding},
  author={Song, Enxin and Chai, Wenhao and Wang, Guanhong and Zhang, Yucheng and Zhou, Haoyang and Wu, Feiyang and Chi, Haozhe and Guo, Xun and Ye, Tian and Zhang, Yanting and others},
  booktitle={Proceedings of the IEEE/CVF Conference on Computer Vision and Pattern Recognition},
  pages={18221--18232},
  year={2024}
}

@inproceedings{maaz2024video,
  title={Video-chatgpt: Towards detailed video understanding via large vision and language models},
  author={Maaz, Muhammad and Rasheed, Hanoona and Khan, Salman and Khan, Fahad},
  booktitle={Proceedings of the 62nd Annual Meeting of the Association for Computational Linguistics (Volume 1: Long Papers)},
  pages={12585--12602},
  year={2024}
}

@article{xu2024slowfast,
  title={{SlowFast-LLaVA}: A strong training-free baseline for video large language models},
  author={Xu, Mingze and Gao, Mingfei and Gan, Zhe and Chen, Hong-You and Lai, Zhengfeng and Gang, Haiming and Kang, Kai and Dehghan, Afshin},
  journal={arXiv:2407.15841},
  year={2024}
}

@article{xu2024pllava,
  title={{PLLaVA}: Parameter-free llava extension from images to videos for video dense captioning},
  author={Xu, Lin and Zhao, Yilin and Zhou, Daquan and Lin, Zhijie and Ng, See Kiong and Feng, Jiashi},
  journal={arXiv:2404.16994},
  year={2024}
}

@article{zhang2024long,
  title={Long Context Transfer from Language to Vision},
  author={Zhang, Peiyuan and Zhang, Kaichen and Li, Bo and Zeng, Guangtao and Yang, Jingkang and Zhang, Yuanhan and Wang, Ziyue and Tan, Haoran and Li, Chunyuan and Liu, Ziwei},
  journal={arXiv:2406.16852},
  year={2024}
}

@article{wang2024qwen2,
  title={{Qwen2-VL}: Enhancing vision-language model's perception of the world at any resolution},
  author={Wang, Peng and Bai, Shuai and Tan, Sinan and Wang, Shijie and Fan, Zhihao and Bai, Jinze and Chen, Keqin and Liu, Xuejing and Wang, Jialin and Ge, Wenbin and others},
  journal={arXiv:2409.12191},
  year={2024}
}

@article{team2023gemini,
  title={Gemini: a family of highly capable multimodal models},
  author={Team, Gemini and Anil, Rohan and Borgeaud, Sebastian and Wu, Yonghui and Alayrac, Jean-Baptiste and Yu, Jiahui and Soricut, Radu and Schalkwyk, Johan and Dai, Andrew M and Hauth, Anja and others},
  journal={arXiv:2312.11805},
  year={2023}
}

@article{chen2025eagle,
  title={Eagle 2.5: Boosting long-context post-training for frontier vision-language models},
  author={Chen, Guo and Li, Zhiqi and Wang, Shihao and Jiang, Jindong and Liu, Yicheng and Lu, Lidong and Huang, De-An and Byeon, Wonmin and Le, Matthieu and Rintamaki, Tuomas and others},
  journal={arXiv preprint arXiv:2504.15271},
  year={2025}
}

@article{jin2024video,
  title={{Video-LaVIT}: Unified video-language pre-training with decoupled visual-motional tokenization},
  author={Jin, Yang and Sun, Zhicheng and Xu, Kun and Chen, Liwei and Jiang, Hao and Huang, Quzhe and Song, Chengru and Liu, Yuliang and Zhang, Di and Song, Yang and others},
  journal={arXiv:2402.03161},
  year={2024}
}

@article{shen2024longvu,
  title = {{LongVU}: Spatiotemporal Adaptive Compression for Long Video-Language Understanding},
  author ={Shen, Xiaoqian and Xiong, Yunyang and Zhao, Changsheng and Wu, Lemeng and Chen, Jun and Zhu, Chenchen and Liu, Zechun and Xiao, Fanyi and Varadarajan, Balakrishnan and Bordes, Florian and Liu, Zhuang and Xu, Hu and J. Kim, Hyunwoo and Soran, Bilge and Krishnamoorthi, Raghuraman and Elhoseiny, Mohamed and Chandra, Vikas},
  journal = {arXiv:2410.17434},
  year = {2024}
}

@inproceedings{agrawal2018don,
  title={Don't just assume; look and answer: Overcoming priors for visual question answering},
  author={Agrawal, Aishwarya and Batra, Dhruv and Parikh, Devi and Kembhavi, Aniruddha},
  booktitle={Proceedings of the IEEE conference on computer vision and pattern recognition},
  pages={4971--4980},
  year={2018}
}

@article{cadene2019rubi,
  title={Rubi: Reducing unimodal biases for visual question answering},
  author={Cadene, Remi and Dancette, Corentin and Cord, Matthieu and Parikh, Devi and others},
  journal={Advances in neural information processing systems},
  volume={32},
  year={2019}
}

@inproceedings{clark2019don,
  title={Don’t take the easy way out: Ensemble based methods for avoiding known dataset biases},
  author={Clark, Christopher and Yatskar, Mark and Zettlemoyer, Luke},
  booktitle={Proceedings of the 2019 conference on empirical methods in natural language processing and the 9th international joint conference on natural language Processing (EMNLP-IJCNLP)},
  pages={4069--4082},
  year={2019}
}

@inproceedings{chen2020counterfactual,
  title={Counterfactual samples synthesizing for robust visual question answering},
  author={Chen, Long and Yan, Xin and Xiao, Jun and Zhang, Hanwang and Pu, Shiliang and Zhuang, Yueting},
  booktitle={Proceedings of the IEEE/CVF conference on computer vision and pattern recognition},
  pages={10800--10809},
  year={2020}
}

@inproceedings{niu2021counterfactual,
  title={Counterfactual vqa: A cause-effect look at language bias},
  author={Niu, Yulei and Tang, Kaihua and Zhang, Hanwang and Lu, Zhiwu and Hua, Xian-Sheng and Wen, Ji-Rong},
  booktitle={Proceedings of the IEEE/CVF conference on computer vision and pattern recognition},
  pages={12700--12710},
  year={2021}
}

@inproceedings{xiao2021next,
  title={Next-qa: Next phase of question-answering to explaining temporal actions},
  author={Xiao, Junbin and Shang, Xindi and Yao, Angela and Chua, Tat-Seng},
  booktitle={Proceedings of the IEEE/CVF conference on computer vision and pattern recognition},
  pages={9777--9786},
  year={2021}
}

@article{wu2024longvideobench,
  title={Longvideobench: A benchmark for long-context interleaved video-language understanding},
  author={Wu, Haoning and Li, Dongxu and Chen, Bei and Li, Junnan},
  journal={Advances in Neural Information Processing Systems},
  volume={37},
  pages={28828--28857},
  year={2024}
}

@article{li2024videochat,
  title={Videochat-flash: Hierarchical compression for long-context video modeling},
  author={Li, Xinhao and Wang, Yi and Yu, Jiashuo and Zeng, Xiangyu and Zhu, Yuhan and Huang, Haian and Gao, Jianfei and Li, Kunchang and He, Yinan and Wang, Chenting and others},
  journal={arXiv preprint arXiv:2501.00574},
  year={2024}
}

@article{liu2025towards,
  title={Towards training-free long video understanding: methods, benchmarks, and open challenges},
  author={Liu, Jingren and Wang, Yun and Zhang, Long and Wang, Yiheng and Xu, Shuning and Wang, Ling and Yan, Jiaqi and Zhang, Dell and Chen, Xiangyu},
  journal={Vicinagearth},
  volume={2},
  number={1},
  pages={6},
  year={2025},
  publisher={Springer}
}

@inproceedings{cheng2025enhancing,
  title={Enhancing long video understanding via hierarchical event-based memory},
  author={Cheng, Dingxin and Li, Mingda and Liu, Jingyu and Guo, Yongxin and Jiang, Bin and Liu, Qingbin and Chen, Xi and Zhao, Bo},
  booktitle={2025 IEEE International Conference on Multimedia and Expo (ICME)},
  pages={1--6},
  year={2025},
  organization={IEEE}
}

@inproceedings{santos2025video,
  title={{$\infty$-Video}: A Training-Free Approach to Long Video Understanding via Continuous-Time Memory Consolidation},
  author={Santos, Saul and Farinhas, Ant{\'o}nio and McNamee, Daniel C. and Martins, Andr{\'e}},
  booktitle={International Conference on Machine Learning},
  pages={52877--52893},
  year={2025},
  organization={PMLR}
}

@inproceedings{wang2025videotree,
  title={Videotree: Adaptive tree-based video representation for llm reasoning on long videos},
  author={Wang, Ziyang and Yu, Shoubin and Stengel-Eskin, Elias and Yoon, Jaehong and Cheng, Feng and Bertasius, Gedas and Bansal, Mohit},
  booktitle={Proceedings of the Computer Vision and Pattern Recognition Conference},
  pages={3272--3283},
  year={2025}
}

@inproceedings{wang2024videoagent,
  title={Videoagent: Long-form video understanding with large language model as agent},
  author={Wang, Xiaohan and Zhang, Yuhui and Zohar, Orr and Yeung-Levy, Serena},
  booktitle={European Conference on Computer Vision},
  pages={58--76},
  year={2024},
  organization={Springer}
}

@inproceedings{chen2025lvagent,
  title={Lvagent: Long video understanding by multi-round dynamical collaboration of mllm agents},
  author={Chen, Boyu and Yue, Zhengrong and Chen, Siran and Wang, Zikang and Liu, Yang and Li, Peng and Wang, Yali},
  booktitle={Proceedings of the IEEE/CVF International Conference on Computer Vision},
  pages={20237--20246},
  year={2025}
}

@article{zhi2025videoagent2,
  title={Videoagent2: Enhancing the llm-based agent system for long-form video understanding by uncertainty-aware cot},
  author={Zhi, Zhuo and Wu, Qiangqiang and Li, Wenbo and Li, Yinchuan and Shao, Kun and Zhou, Kaiwen and others},
  journal={arXiv preprint arXiv:2504.04471},
  year={2025}
}

@inproceedings{wang2026videochat,
  title={Videochat-a1: Thinking with long videos by chain-of-shot reasoning},
  author={Wang, Zikang and Chen, Boyu and Yue, Zhengrong and Wang, Yi and Qiao, Yu and Wang, Limin and Wang, Yali},
  booktitle={Proceedings of the AAAI Conference on Artificial Intelligence},
  volume={40},
  number={13},
  pages={10467--10475},
  year={2026}
}

@inproceedings{tang2025adaptive,
  title={Adaptive keyframe sampling for long video understanding},
  author={Tang, Xi and Qiu, Jihao and Xie, Lingxi and Tian, Yunjie and Jiao, Jianbin and Ye, Qixiang},
  booktitle={Proceedings of the Computer Vision and Pattern Recognition Conference},
  pages={29118--29128},
  year={2025}
}

@inproceedings{zhang2025beyond,
  title={Beyond training: Dynamic token merging for zero-shot video understanding},
  author={Zhang, Yiming and Zhao, Zhuokai and Chen, Zhaorun and Ding, Zenghui and Yang, Xianjun and Sun, Yining},
  booktitle={Proceedings of the IEEE/CVF International Conference on Computer Vision},
  pages={22046--22055},
  year={2025}
}

@article{yu2024frame,
  title={Frame-voyager: Learning to query frames for video large language models},
  author={Yu, Sicheng and Jin, Chengkai and Wang, Huanyu and Chen, Zhenghao and Jin, Sheng and Zuo, Zhongrong and Xu, Xiaolei and Sun, Zhenbang and Zhang, Bingni and Wu, Jiawei and others},
  journal={arXiv preprint arXiv:2410.03226},
  year={2024}
}

@article{zhang2025flexselect,
  title={Flexselect: Flexible token selection for efficient long video understanding},
  author={Zhang, Yunzhu and Lu, Yu and Wang, Tianyi and Rao, Fengyun and Yang, Yi and Zhu, Linchao},
  journal={arXiv preprint arXiv:2506.00993},
  year={2025}
}

@article{chen2024sharegpt4video,
  title={Sharegpt4video: Improving video understanding and generation with better captions},
  author={Chen, Lin and Wei, Xilin and Li, Jinsong and Dong, Xiaoyi and Zhang, Pan and Zang, Yuhang and Chen, Zehui and Duan, Haodong and Lin, Bin and Tang, Zhenyu and others},
  journal={Advances in Neural Information Processing Systems},
  volume={37},
  pages={19472--19495},
  year={2024}
}

@article{zhang2024llava,
  title={Llava-video: Video instruction tuning with synthetic data},
  author={Zhang, Yuanhan and Wu, Jinming and Li, Wei and Li, Bo and Ma, Zejun and Liu, Ziwei and Li, Chunyuan},
  journal={arXiv preprint arXiv:2410.02713},
  year={2024}
}

@inproceedings{fu2025video,
  title={Video-mme: The first-ever comprehensive evaluation benchmark of multi-modal llms in video analysis},
  author={Fu, Chaoyou and Dai, Yuhan and Luo, Yongdong and Li, Lei and Ren, Shuhuai and Zhang, Renrui and Wang, Zihan and Zhou, Chenyu and Shen, Yunhang and Zhang, Mengdan and others},
  booktitle={Proceedings of the IEEE/CVF conference on computer vision and pattern recognition},
  pages={24108--24118},
  year={2025}
}

@article{zhou2024mlvu,
  title={Mlvu: A comprehensive benchmark for multi-task long video understanding},
  author={Zhou, Junjie and Shu, Yan and Zhao, Bo and Wu, Boya and Xiao, Shitao and Yang, Xi and Xiong, Yongping and Zhang, Bo and Huang, Tiejun and Liu, Zheng},
  journal={arXiv preprint arXiv:2406.04264},
  volume={2},
  number={5},
  pages={6},
  year={2024}
}

@inproceedings{shu2025video,
  title={Video-xl: Extra-long vision language model for hour-scale video understanding},
  author={Shu, Yan and Liu, Zheng and Zhang, Peitian and Qin, Minghao and Zhou, Junjie and Liang, Zhengyang and Huang, Tiejun and Zhao, Bo},
  booktitle={Proceedings of the Computer Vision and Pattern Recognition Conference},
  pages={26160--26169},
  year={2025}
}

@inproceedings{liu2025nvila,
  title={Nvila: Efficient frontier visual language models},
  author={Liu, Zhijian and Zhu, Ligeng and Shi, Baifeng and Zhang, Zhuoyang and Lou, Yuming and Yang, Shang and Xi, Haocheng and Cao, Shiyi and Gu, Yuxian and Li, Dacheng and others},
  booktitle={Proceedings of the IEEE/CVF Conference on Computer Vision and Pattern Recognition},
  pages={4122--4134},
  year={2025}
}

@article{zhang2025videollama,
  title={Videollama 3: Frontier multimodal foundation models for image and video understanding},
  author={Zhang, Boqiang and Li, Kehan and Cheng, Zesen and Hu, Zhiqiang and Yuan, Yuqian and Chen, Guanzheng and Leng, Sicong and Jiang, Yuming and Zhang, Hang and Li, Xin and others},
  journal={arXiv preprint arXiv:2501.13106},
  year={2025}
}

@misc{gpt4o,
  author = {{OpenAI}},
  title = {Hello {GPT}-4o},
  year = {2024},
  howpublished = {\url{https://openai.com/index/hello-gpt-4o/}},
  note = {Accessed: 2024-05-20}
}

@article{cheng2025scaling,
  title={Scaling video-language models to 10k frames via hierarchical differential distillation},
  author={Cheng, Chuanqi and Guan, Jian and Wu, Wei and Yan, Rui},
  journal={arXiv preprint arXiv:2504.02438},
  year={2025}
}

@article{zhu2025internvl3,
  title={Internvl3: Exploring advanced training and test-time recipes for open-source multimodal models},
  author={Zhu, Jinguo and Wang, Weiyun and Chen, Zhe and Liu, Zhaoyang and Ye, Shenglong and Gu, Lixin and Tian, Hao and Duan, Yuchen and Su, Weijie and Shao, Jie and others},
  journal={arXiv preprint arXiv:2504.10479},
  year={2025}
}

@article{bai2025qwen3,
  title={Qwen3-vl technical report},
  author={Bai, Shuai and Cai, Yuxuan and Chen, Ruizhe and Chen, Keqin and Chen, Xionghui and Cheng, Zesen and Deng, Lianghao and Ding, Wei and Gao, Chang and Ge, Chunjiang and others},
  journal={arXiv preprint arXiv:2511.21631},
  year={2025}
}

@article{ye2024mplug,
  title={mplug-owl3: Towards long image-sequence understanding in multi-modal large language models},
  author={Ye, Jiabo and Xu, Haiyang and Liu, Haowei and Hu, Anwen and Yan, Ming and Qian, Qi and Zhang, Ji and Huang, Fei and Zhou, Jingren},
  journal={arXiv preprint arXiv:2408.04840},
  year={2024}
}

@article{ma2025videoeval,
  title={Videoeval-pro: Robust and realistic long video understanding evaluation},
  author={Ma, Wentao and Ren, Weiming and Jia, Yiming and Li, Zhuofeng and Nie, Ping and Zhang, Ge and Chen, Wenhu},
  journal={arXiv preprint arXiv:2505.14640},
  year={2025}
}

@inproceedings{li2024knn,
  title={KNN Transformer with Pyramid Prompts for Few-Shot Learning},
  author={Li, Wenhao and Wang, Qiangchang and Zhao, Peng and Yin, Yilong},
  booktitle={Proceedings of the 32nd ACM International Conference on Multimedia},
  pages={1082--1091},
  year={2024}
}

@article{li2026semmsa,
  title={SemMSA: Latent Semantic-Aided Robust Multimodal Sentiment Analysis with Incomplete Data},
  author={Li, Wenhao and Wu, Zhibin and Xiao, Chong and Wang, Qiangchang},
  journal={arXiv preprint arXiv:2609.30238},
  year={2026}
}

@inproceedings{li2025vtfsl,
  title={VT-FSL: Bridging Vision and Text with LLMs for Few-Shot Learning},
  author={Li, Wenhao and Wang, Qiangchang and Meng, Xianjing and Wu, Zhibin and Yin, Yilong},
  booktitle={NeurIPS},
  volume={38},
  pages={113697--113727},
  year={2025}
}

@inproceedings{li2026dvla,
  title={DVLA-RL: Dual-Level Vision-Language Alignment with Reinforcement Learning Gating for Few-Shot Learning},
  author={Li, Wenhao and Meng, Xianjing and Wang, Qiangchang and Han, Zhongyi and Wu, Zhibin and Yin, Yilong},
 booktitle={ICLR},
 year={2026}
}

@inproceedings{liang2026multimodal,
  title={Multimodal Causality-Driven Representation Learning for Generalizable Medical Image Segmentation},
  author={Liang, Xusheng and Zhou, Lihua and Li, Nianxin and Xu, Miao and Song, Ziyang and Yi, Dong and Wu, Jinlin and Ma, Jiawei and Liu, Hongbin and Lei, Zhen and others},
  booktitle={Proceedings of the IEEE/CVF Conference on Computer Vision and Pattern Recognition},
  pages={13670--13679},
  year={2026}
}

@inproceedings{pu2026f,
  title={{$F^2$-Assist}: Multi-Phase Fetal Growth Forecast and Report Generation from Ultrasound Examination},
  author={Pu, Bin and Liang, Xusheng and Ding, Xinpeng and Wu, Jinlin and Lei, Zhen and Li, Shengli and Li, Kenli and Ma, Jiawei},
  booktitle={Proceedings of the IEEE/CVF Conference on Computer Vision and Pattern Recognition},
  pages={35341--35350},
  year={2026}
}

@inproceedings{wu2024surgbox,
  title={Surgbox: Agent-driven operating room sandbox with surgery copilot},
  author={Wu, Jinlin and Liang, Xusheng and Bai, Xuexue and Chen, Zhen},
  booktitle={2024 IEEE International Conference on Big Data (BigData)},
  pages={2041--2048},
  year={2024},
  organization={IEEE}
}

@inproceedings{ma2023digeo,
  title={Digeo: Discriminative geometry-aware learning for generalized few-shot object detection},
  author={Ma, Jiawei and Niu, Yulei and Xu, Jincheng and Huang, Shiyuan and Han, Guangxing and Chang, Shih-Fu},
  booktitle={2023 IEEE/CVF Conference on Computer Vision and Pattern Recognition (CVPR)},
  pages={3208--3218},
  year={2023},
  organization={IEEE}
}

@inproceedings{ma2024mode,
  title={Mode: Clip data experts via clustering},
  author={Ma, Jiawei and Huang, Po-Yao and Xie, Saining and Li, Shang-Wen and Zettlemoyer, Luke and Chang, Shih-Fu and Yih, Wen-Tau and Xu, Hu},
  booktitle={2024 IEEE/CVF Conference on Computer Vision and Pattern Recognition (CVPR)},
  pages={26344--26353},
  year={2024},
  organization={IEEE}
}

\clearpage 
\appendix

\section{Technical appendices and supplementary material}

\subsection{Limitation}
\label{sec:limitation}
CLEAR is training-free but not cost-free. Since it uses question-only and
salient-view probes to estimate shortcut residuals, each sample requires
additional auxiliary forward passes before final full-video decoding. This
multi-perspective inference introduces extra latency and computation compared
with a vanilla single-pass baseline, particularly for long videos and large
MLLM backbones.

Our current focus is to validate the effectiveness of controlled shortcut
probes for evidence-grounded reasoning. Improving the efficiency of CLEAR is an
important future direction. Potential solutions include adaptive probe
activation, lightweight probe approximations, representation caching, shared
early-layer computation, and early exiting when the discrepancy between the
full-video and shortcut-probe trajectories is small.

\subsection{Implementation Details}
\label{sec:implementation_details}

All experiments are conducted with frozen open-source MLLMs, including LLaVA-Video-7B, InternVL3-8B, and Qwen3-VL-8B. 
Closed-source models are only reported as reference systems since their hidden states are inaccessible. 
Unless otherwise specified, we uniformly sample 64 frames from each video for the final full-video inference path. 
For CLEAR, we run two auxiliary probes before final decoding: a question-only probe and a salient-view probe. 
The question-only probe removes all visual inputs, while the salient-view probe selects the top-$K$ query-relevant clips according to attention scores from query tokens to visual tokens; we use $K=4$ by default. 
The probes are only used to compute shortcut residuals and never directly generate the final answer. 
For each view, hidden states are extracted at the same textual decision position.

Residual steering is applied to late decoder layers. 
For Qwen3-VL-8B, we use layers 28--35 by default, and for other backbones we use the corresponding late-layer block with the same relative depth. 
A probe-specific residual is activated only when $d_l^m=1-\cos(h_l^f,h_l^m)>\tau$, with $\tau=0.15$ for all experiments. 
The default residual strength is $\alpha=0.15$, selected from $\{0.05,0.10,0.15,0.25,0.35\}$ according to the accuracy--shortcut suppression trade-off. 

All experiments are run on NVIDIA H100 GPUs with batch size one for long-video inference. 
Efficiency is reported as relative inference time normalized by the vanilla 64-frame baseline of the same backbone, including auxiliary probing, residual construction, and final full-video decoding.

\paragraph{Open-ended video QA.}
To examine whether CLEAR is limited to multiple-choice prediction, we further evaluate it on VideoEvalPro, an open-ended video question answering benchmark. As shown in Table~\ref{tab:videoevalpro}, CLEAR consistently improves free-form generation across all three backbones, with gains of +1.2, +1.5, and +3.4 points on LLaVA-Video-7B, InternVL3-8B, and Qwen3-VL-8B, respectively. Since VideoEvalPro requires models to generate answers without candidate options, these improvements suggest that CLEAR does not merely calibrate multiple-choice scores, but also steers generation toward more evidence-consistent responses.

\begin{table}[h]
\centering
\caption{
\textbf{Open-ended evaluation on VideoEvalPro\cite{ma2025videoeval}.}
CLEAR improves free-form video question answering, suggesting that its gains are not limited to multiple-choice task.
}
\label{tab:videoevalpro}
\small
\begin{tabular}{lccc}
\toprule
Model & Frames & Method & VideoEvalPro \\
\midrule
LLaVA-Video-7B & 64 & Baseline & 22.8 \\
LLaVA-Video-7B & 64 & + CLEAR & 24.0 \\
InternVL3-8B & 64 & Baseline & 23.8 \\
InternVL3-8B & 64 & + CLEAR & 25.3 \\
Qwen3-VL-8B & 64 & Baseline & 25.2 \\
Qwen3-VL-8B & 64 & + CLEAR & 28.6 \\
\bottomrule
\end{tabular}
\end{table}

\paragraph{Qualitative analysis.}
Figure~\ref{fig:open_qa} shows an open-ended QA example where CLEAR corrects a shortcut-biased response and produces an answer better aligned with the decisive temporal evidence.

\begin{figure}[t] \centering \includegraphics[width=\linewidth]{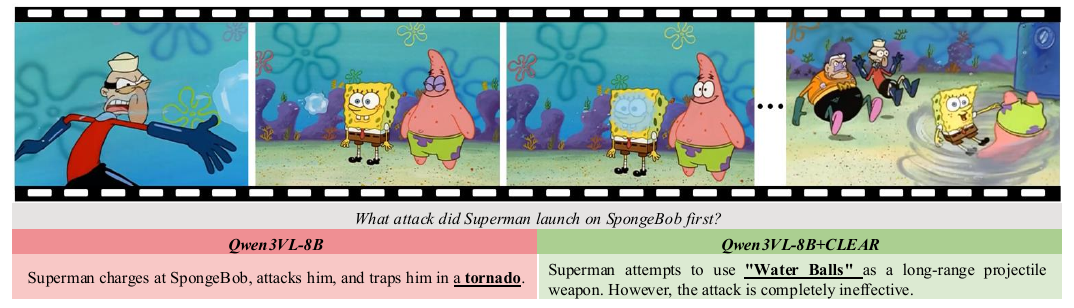} \caption{Qualitative effect of CLEAR on open-ended video QA. } \label{fig:open_qa} \end{figure}

\subsection{Broader Impact}
\label{sec:social_impacts}
CLEAR operates as a training-free inference-time module that steers long-video MLLMs toward evidence-consistent predictions. It may improve reliability in long-form video understanding tasks, but it cannot prevent the base MLLMs from generating erroneous, biased, or harmful responses. Therefore, CLEAR should be viewed as a complementary robustness technique rather than a standalone safety mechanism, especially in high-stakes applications.

\clearpage 
\section*{NeurIPS Paper Checklist}

\begin{enumerate}

\item {\bf Claims}
    \item[] Question: Do the main claims made in the abstract and introduction accurately reflect the paper's contributions and scope?
    \item[] Answer: \answerYes{} 
    \item[] Justification: We carefully described our contributions in the abstract and introduction.

\item {\bf Limitations}
    \item[] Question: Does the paper discuss the limitations of the work performed by the authors?
    \item[] Answer: \answerYes{} 
    \item[] Justification: We discussed our limitations in appendix \ref{sec:limitation}

\item {\bf Theory assumptions and proofs}
    \item[] Question: For each theoretical result, does the paper provide the full set of assumptions and a complete (and correct) proof?
    \item[] Answer: \answerNA{} 
    \item[] Justification: This paper is not about theory.

    \item {\bf Experimental result reproducibility}
    \item[] Question: Does the paper fully disclose all the information needed to reproduce the main experimental results of the paper to the extent that it affects the main claims and/or conclusions of the paper (regardless of whether the code and data are provided or not)?
    \item[] Answer: \answerYes{} 
    \item[] Justification: We provided details about our methods and implementation in the main paper and appendix. We will also upload our main code as supplementary materials.

\item {\bf Open access to data and code}
    \item[] Question: Does the paper provide open access to the data and code, with sufficient instructions to faithfully reproduce the main experimental results, as described in supplemental material?
    \item[] Answer: \answerYes{} 
    \item[] Justification: We plan to release the CLEAR scripts, S-MME annotations and evaluation scripts

\item {\bf Experimental setting/details}
    \item[] Question: Does the paper specify all the training and test details (e.g., data splits, hyperparameters, how they were chosen, type of optimizer) necessary to understand the results?
    \item[] Answer: \answerYes{} 
    \item[] Justification: We present the experimental setup and details in the main paper

\item {\bf Experiment statistical significance}
    \item[] Question: Does the paper report error bars suitably and correctly defined or other appropriate information about the statistical significance of the experiments?
    \item[] Answer: \answerNo{} 
    \item[] Justification: However, we provide our code, detail configuration and random seeds to facilitate the reproducibility of our results.


\item {\bf Experiments compute resources}
    \item[] Question: For each experiment, does the paper provide sufficient information on the computer resources (type of compute workers, memory, time of execution) needed to reproduce the experiments?
    \item[] Answer: \answerYes{} 
    \item[] Justification: We clarify our computation resources at implementation details in appendix.
    
\item {\bf Code of ethics}
    \item[] Question: Does the research conducted in the paper conform, in every respect, with the NeurIPS Code of Ethics \url{https://neurips.cc/public/EthicsGuidelines}?
    \item[] Answer: \answerYes{} 
    \item[] Justification: We carefully reviewed the NeurIPS Code of Ethics.

\item {\bf Broader impacts}
    \item[] Question: Does the paper discuss both potential positive societal impacts and negative societal impacts of the work performed?
    \item[] Answer: \answerYes{} 
    \item[] Justification: We discussed societal impact in appendix \ref{sec:social_impacts}.
    
\item {\bf Safeguards}
    \item[] Question: Does the paper describe safeguards that have been put in place for responsible release of data or models that have a high risk for misuse (e.g., pre-trained language models, image generators, or scraped datasets)?
    \item[] Answer: \answerYes{} 
    \item[] Justification: The data and models used in this paper have been extensively used in the multi-modal understanding community, and have undergone comprehensive safety risk assessments.

\item {\bf Licenses for existing assets}
    \item[] Question: Are the creators or original owners of assets (e.g., code, data, models), used in the paper, properly credited and are the license and terms of use explicitly mentioned and properly respected?
    \item[] Answer: \answerYes{} 
    \item[] Justification: We provided appropriate citations in the reference section.

\item {\bf New assets}
    \item[] Question: Are new assets introduced in the paper well documented and is the documentation provided alongside the assets?
    \item[] Answer: \answerYes{} 
    \item[] Justification: We have included the code and benchmark.

\item {\bf Crowdsourcing and research with human subjects}
    \item[] Question: For crowdsourcing experiments and research with human subjects, does the paper include the full text of instructions given to participants and screenshots, if applicable, as well as details about compensation (if any)? 
    \item[] Answer: \answerNA{} 
    \item[] Justification: The paper does not conduct human-subject experiments. Human involvement is limited to internal annotation and verification of video-QA samples; no personal data, behavioral measurements, or user studies are collected.

\item {\bf Institutional review board (IRB) approvals or equivalent for research with human subjects}
    \item[] Question: Does the paper describe potential risks incurred by study participants, whether such risks were disclosed to the subjects, and whether Institutional Review Board (IRB) approvals (or an equivalent approval/review based on the requirements of your country or institution) were obtained?
    \item[] Answer: \answerNA{} 
    \item[] Justification: The paper does not involve human-subject experiments. The human annotation process is used only for dataset verification and does not collect personal or sensitive information from annotators.

\item {\bf Declaration of LLM usage}
    \item[] Question: Does the paper describe the usage of LLMs if it is an important, original, or non-standard component of the core methods in this research? Note that if the LLM is used only for writing, editing, or formatting purposes and does \emph{not} impact the core methodology, scientific rigor, or originality of the research, declaration is not required.
    \item[] Answer: \answerNA{} 
    \item[] Justification: The core method developed in this study was derived through extensive
hypothesis-experimentation cycles, without utilizing LLMs as any essential, novel, or
non-standard components.

\end{enumerate}

\end{document}